\documentclass[11pt]{article}
\usepackage{amsmath}
\usepackage{amssymb}
\usepackage{amsthm}
\usepackage{graphicx}
\usepackage{float}
\usepackage{booktabs}
\usepackage{siunitx}
\usepackage{longtable}
\usepackage[T1]{fontenc}
\usepackage[utf8]{inputenc}
\usepackage[authoryear]{natbib}
\usepackage[margin=1in]{geometry}
\usepackage{microtype}
\usepackage[hidelinks]{hyperref}

\title{A Dual-Track Curation-and-Classification Framework\\ for Resolving Ground-Truth Label Noise\\ in Operational Sentinel-2 Wheat Area Estimation}

\author{
Kasimali Agharia \and 
Ujjwal Kumar Gupta \\
\vspace{0.5em} 
\small Space Applications Centre (SAC), Indian Space Research Organisation (ISRO)
}
\date{\today}

\begin{document}

\maketitle

\begin{abstract}
Operational estimation of wheat-cultivated area is persistently constrained by discordance between administrative record-keeping and remotely sensed classification products, an inconsistency this study terms the \emph{administrative reference discordance}. We address this discordance for the 2022 Rabi season, with classifier deployment and area reconciliation in Patiala district, Punjab, India, using a thirteen-timestep Sentinel-2 NDVI time series (15 October to 15 April, approximately 15-day compositing interval) smoothed via Savitzky--Golay filtering (window $=5$, polynomial order $=2$). A curated 849-sample reference dataset drawn from across Punjab State, developed through an iterative, rule-based bootstrapping procedure across three successive sample batches, underpins both the sensitivity analysis and the operational classifier reported below. Feature sensitivity across the thirteen phenological timesteps was independently assessed using both the standardised Cohen's $d$ effect size (wheat versus non-wheat, $n=849$) and gradient-boosted information gain; the two measures converge on the February-to-March grain-fill window as most discriminative (peak $|d|=2.21$ at 01 March) while diverging on which edge timesteps contribute least -- a discrepancy reported rather than resolved in favour of either single metric. Four classifiers -- a one-dimensional convolutional network, a stacked long short-term memory (LSTM) network, a hybrid CNN--LSTM, and an Extreme Gradient Boosting (XGBoost) ensemble -- were benchmarked on an identical stratified 679/170 sample split; XGBoost achieved the highest overall accuracy (78.82\%) against substantially lower deep-learning baselines (66.47\%, 64.12\%, and 64.71\% respectively), consistent with tree-based ensembles' favourable parameter-to-sample ratio in low-sample operational regimes ($n=849$). At full-population deployment across 36{,}255{,}347 valid district pixels, the operational classifier attained 86.31\% precision, 71.05\% recall, an F1-score of 0.7794, and a Kappa coefficient of 0.1882, which reflects class prevalence at population scale together with residual confusion within the non-wheat aggregate. The operational classifier's predicted wheat extent (240{,}682.51\,ha) deviated by only $+2.99\%$ from the official government tabular target (233{,}700.00\,ha), whereas the government's own spatial reference mask exhibited a $+25.11\%$ positive area bias (292{,}386.91\,ha) against the identical target. This asymmetry indicates that a classifier trained exclusively on an auditor-curated, noise-filtered reference set can reconcile more closely with the official tabular area than the administrative spatial product conventionally used to validate it, in the one district and season studied. We present this dual-track curation-and-classification framework, together with its documented statistical limitations, as a methodological reference for crop-area reconciliation in label-noisy administrative settings.
\end{abstract}

\vspace{1em}

\newpage
\section{Introduction}
\label{sec:intro}

Remote sensing has become the de facto instrument for large-area, sub-seasonal crop monitoring, yet operational adoption is repeatedly undermined by a problem that receives comparatively little attention in the classification literature: the reference data against which classifiers are trained and validated are themselves imperfect. Administrative crop statistics, compiled through tabular reporting mechanisms, and spatial reference masks, digitised or interpolated from field boundary records, frequently diverge from one another by a non-trivial margin even when both nominally describe the same growing season and administrative unit. Where a remotely sensed classification product is validated against a spatial mask that itself carries systematic bias, the resulting accuracy assessment inherits that bias -- a circularity this study refers to as the \emph{administrative reference discordance}. The discordance is not merely definitional: as shown in Section~\ref{sec:results-forward}, a government spatial mask for Patiala district's 2022 Rabi wheat extent diverges from the corresponding official tabular statistic by $+25.11\%$, a magnitude sufficient to materially distort any accuracy assessment that treats the spatial mask as unconditionally authoritative. This problem exists against a backdrop of documented regional success in wheat-specific remote sensing for this agro-climatic zone. \citet{mohite2019}, working across districts of Punjab and Haryana, demonstrated early-season wheat area mapping accuracies of up to 88.31\% using a Random Forest classifier on fused Sentinel-1 SAR and Sentinel-2 optical features, with NDVI alone achieving a closely comparable 87.19\%  and outperforming the SAR-only configuration (79.16\%). This regional precedent motivates the present study's reliance on an NDVI-only feature space (Section~\ref{sec:xgboost}) while highlighting that classification accuracy in isolation, as established by that prior work, does not address the area-estimation reconciliation problem this study investigates.

A second, related constraint concerns the choice of classification architecture under realistic operational sample budgets. Deep learning architectures -- convolutional and recurrent networks in particular -- have demonstrated strong performance in crop-type mapping where labelled training samples number in the thousands to tens of thousands. Operational settings requiring hand-verified, agronomically defensible ground reference, however, routinely yield far smaller curated sample sets. Under such constraints, highly parameterised sequence models are prone to overfitting the idiosyncrasies of a small training partition rather than learning generalisable phenological structure, whereas shallow, axis-aligned decision ensembles operating on a small number of discrete, phenologically meaningful features have a more favourable parameter-to-sample ratio and correspondingly better-behaved generalisation properties \citep{chen2016xgboost}. This study treats that architectural trade-off as an empirical question rather than an assumption, benchmarking four classifier families under an identical, small, iteratively curated sample budget ($n=849$).

This study makes three contributions. First, it introduces a dual-track architecture that explicitly separates the diagnostic, curation-stage function of validating and de-noising ground-reference samples (Track 2) from the operational, deployment-stage function of pixel-wise classification across the full district raster (Track 1), rather than conflating the two as a single end-to-end pipeline; Track 2 itself is realised as an iterative bootstrapping and rule-refinement procedure across three successive sample batches, rather than a single curation pass. Second, it applies a stratified sensitivity analysis -- combining an independent, distribution-based effect size (Cohen's $d$) with model-internal information gain -- to the thirteen-timestep NDVI feature space, reporting both tails of the resulting ranking rather than only the most discriminative features. Third, it quantifies the administrative reference discordance for the study district, showing that an operational classifier trained on a small, auditor-curated sample set can reconcile more closely with official administrative statistics than the spatial reference product conventionally used for its own validation.

The remainder of this paper is organised as follows. Section~\ref{sec:studyarea} describes the study area, satellite data sources, and preprocessing pipeline, including a disclosed limitation regarding spatial autocorrelation in the sampling design. Section~\ref{sec:methods} details the dual-track methodology: the Track 2 diagnostic auditor's iterative bootstrapping procedure, and the Track 1 operational XGBoost classifier deployed across the full study district.

\section{Study Area and Dataset}
\label{sec:studyarea}

\subsection{Study area and growing season}
The classifier was deployed, and its wheat-area estimate reconciled with administrative statistics, in Patiala district, Punjab, India, an intensively cultivated agricultural district in the Indo-Gangetic plain characterised by a rice--wheat rotation system; the reference samples used for training were drawn from across Punjab State. Analysis is restricted to the 2022 Rabi (winter) cropping season, spanning sowing from approximately October 2021 through harvest in April 2022 \citep{pau_rabi}. The official administrative tabular wheat-area statistic and the government spatial reference mask evaluated in Section~\ref{sec:acreage} were obtained from the relevant state-level agricultural and remote-sensing authority responsible for maintaining Punjab's crop-area records, accessed via an institutional data-sharing arrangement with the Space Applications Centre (SAC), Indian Space Research Organisation (ISRO), where this research was conducted. Patiala district has itself been the subject of prior multi-source remote-sensing assessment of rabi crop progression and condition \citep{sahay2014}, providing established regional precedent for the wheat- and mustard-focused monitoring approach adopted here.

\subsection{Optical time series construction}
Sentinel-2 Level-2A surface reflectance imagery was filtered to scenes with cloudy-pixel-percentage below 20\%, cloud- and cirrus-masked using the QA60 bitmask (bits 10 and 11), and reduced to the Normalized Difference Vegetation Index (NDVI). Cloud-masked daily mosaics were composited into thirteen fixed calendar timesteps at approximately 15-day intervals, spanning 15 October ($t_{1}$) through 15 April ($t_{13}$):

\begin{center}
\begin{tabular}{ccc|ccc}
\toprule
Timestep & Date & & Timestep & Date \\
\midrule
$t_{1}$ & 15 Oct & & $t_{8}$ & 01 Feb \\
$t_{2}$ & 01 Nov & & $t_{9}$ & 15 Feb \\
$t_{3}$ & 15 Nov & & $t_{10}$ & 01 Mar \\
$t_{4}$ & 01 Dec & & $t_{11}$ & 15 Mar \\
$t_{5}$ & 15 Dec & & $t_{12}$ & 01 Apr \\
$t_{6}$ & 01 Jan & & $t_{13}$ & 15 Apr \\
$t_{7}$ & 15 Jan & & & \\
\bottomrule
\end{tabular}
\end{center}

Residual acquisition noise in the resulting thirteen-point NDVI curve was suppressed using a Savitzky--Golay filter with a five-sample window and second-order polynomial fit \citep{savitzky1964smoothing}, applied independently to each sample location's temporal profile.

\subsection{Sample design and disclosed limitation}
Ground reference points were generated through stratified sampling across Punjab State at the Sentinel-2 pixel scale, seeded from a prior binary wheat/non-wheat institutional classification (Section~\ref{sec:seeding}); no minimum inter-point spacing constraint was enforced during this seeding step. Given the 10\,m sensor resolution and the spatial contiguity of individual agricultural parcels, a subset of sampled points -- particularly those subsequently allocated to different partitions of the train/validation split described in Section~\ref{sec:xgboost} -- may be spatially proximal and therefore non-independent in the strict geostatistical sense. This is disclosed as a limitation on the strict independence of the reported validation accuracies rather than as an invalidation of the reported figures; a spatially blocked resampling scheme is recommended as a direction for future validation of the framework presented here.

\subsection{Curated reference dataset}
The final curated ground-reference dataset comprises 849 samples across twelve phenological classes (wheat -- standard and late/double-sown, with atmospheric-artifact variants of each; mustard; potato and short-Rabi crops; fodder/berseem; forest; barren land; urban; and water), constructed through the iterative dual-track curation procedure described in Section~\ref{sec:methods}.

\section{Methodology}
\label{sec:methods}

\begin{figure}[htbp]
\vspace{-1.5cm}
\centering
\includegraphics[width=0.74\linewidth]{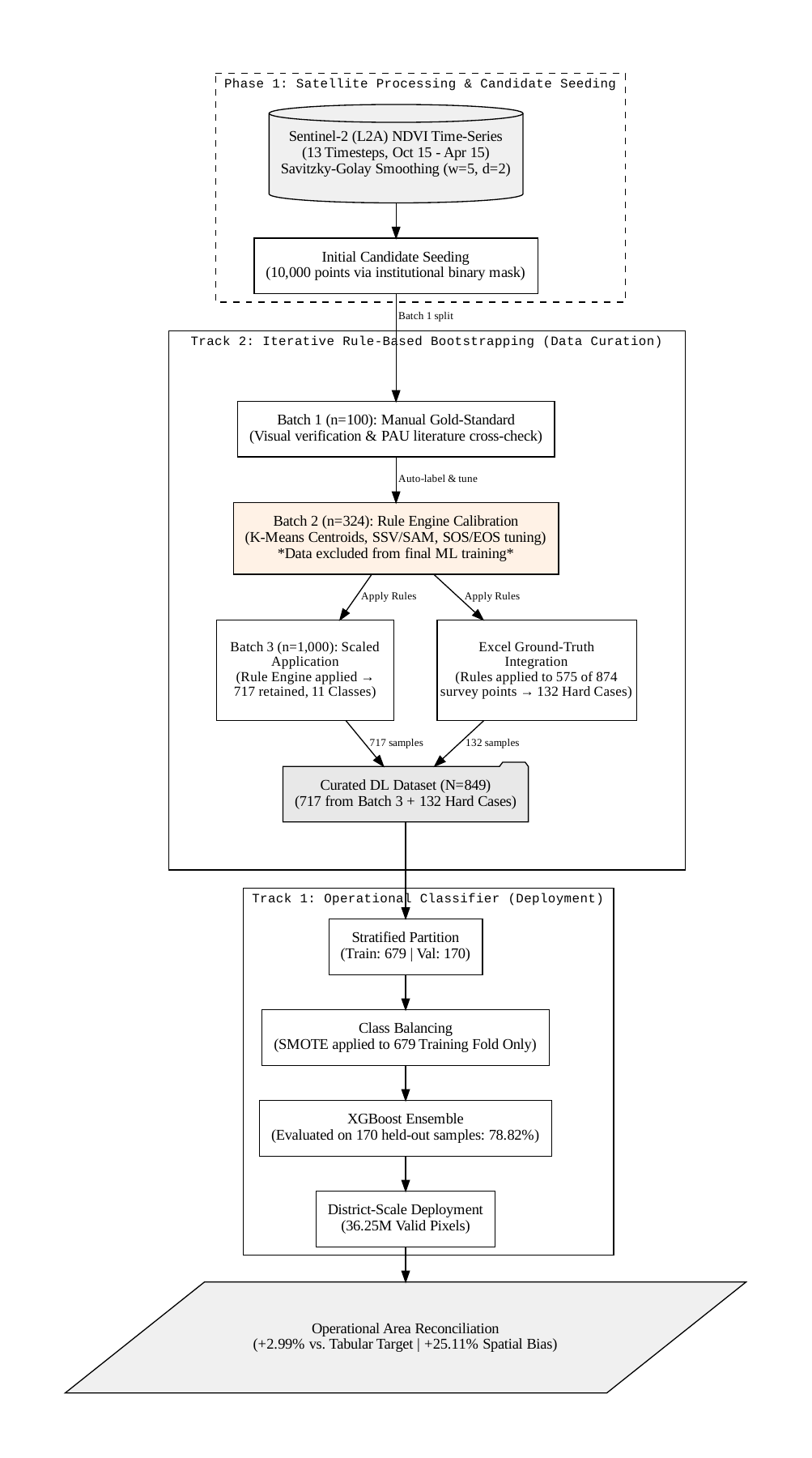}
\caption{Dual-track architecture overview. Track 2 (diagnostic auditor) operates upstream, via the iterative bootstrapping procedure detailed in Section~\ref{sec:methods}, to produce the curated 849-sample training set; Track 1 (operational XGBoost classifier) is trained exclusively on that curated set and deployed pixel-wise across the full study district.}
\label{fig:flowchart}
\end{figure}

The classification framework is organised into two functionally distinct tracks. Track 2, the diagnostic auditor, operates upstream of model training and is responsible for curating a high-fidelity, phenologically validated reference sample set from a larger pool of candidate ground points, through an iterative bootstrapping procedure spanning three successive sample batches (Sections~\ref{sec:seeding}--\ref{sec:finalassembly}). Track 1, the operational classifier, is trained exclusively on the Track 2 output and is the component deployed across the full study district raster (Section~\ref{sec:xgboost}). This separation is deliberate: Track 2's rule-based and distance-metric machinery is well suited to fine-grained, per-sample quality control against a small candidate pool, but was not designed as, and is not evaluated in this paper as, a standalone pixel-wise classifier at deployment scale. Its principal contribution to the framework is therefore curation, not classification. This dual-track division of labour is summarised in Figure~\ref{fig:flowchart}.

\subsection{Initial GEE Seeding}
\label{sec:seeding}

Candidate ground-reference locations were seeded from a prior institutional binary wheat/non-wheat classification asset, using stratified sampling on the asset's own class values to draw candidate points from both wheat- and non-wheat-labelled regions. This initial pool establishes candidate \emph{locations} only; the binary label attached at seeding is not carried forward as a final training label at any subsequent stage of the pipeline. The initial pool comprised 10{,}000 candidate points drawn across Punjab State, of which three successive processing batches of 100, 500, and 1{,}000 points were processed, with the remainder reserved for potential future extension.

\subsection{Batch 1: Manual Gold-Standard Construction}
\label{sec:batch1}

The first batch (100 candidate points) was individually verified through manual visual comparison of each point's Sentinel-2 NDVI temporal profile against the corresponding Google Earth Engine satellite imagery, screening for points falling on roads, buildings, or non-agricultural land inadvertently included by the initial seeding step. The class taxonomy was expanded during this stage beyond a binary wheat/non-wheat scheme to include water, forest, fodder, and other land-cover classes present in the study district. Phenological plausibility was cross-checked against regional Rabi crop growth-stage literature \citep{pau_rabi}, including expected start-of-season, peak, and end-of-season timing. This manually verified Batch 1 set constitutes the first internal gold-standard reference used to bootstrap subsequent, larger batches.

\subsection{Batch 2: Semi-Automated Bootstrapping and Rule-Engine Construction}
\label{sec:batch2}

Batch 2 (500 candidate points) was automatically pre-labelled using distance-based matching against the Batch 1 gold-standard curves, then subjected to the same individual manual review process as Batch 1. This yielded a second, larger gold-standard set (approximately 324 samples) used exclusively to construct the distance-metric and clustering machinery underlying the remainder of the curation pipeline.

\subsubsection{Distance and similarity metrics}
Candidate sample curves were compared against class-representative centroids using six complementary distance and similarity metrics. Given two thirteen-dimensional NDVI vectors $\mathbf{x}$ and $\mathbf{y}$:

\begin{align}
\text{Euclidean distance:} \quad & \mathrm{ED}(\mathbf{x},\mathbf{y}) = \lVert \mathbf{x} - \mathbf{y} \rVert_{2} \\
\text{Shape correlation:} \quad & \mathrm{SCS}(\mathbf{x},\mathbf{y}) = \frac{\mathrm{cov}(\mathbf{x},\mathbf{y})}{\sigma_{\mathbf{x}}\sigma_{\mathbf{y}}} \\
\text{Shape--scale value:} \quad & \mathrm{SSV}(\mathbf{x},\mathbf{y}) = \sqrt{\mathrm{ED}(\mathbf{x},\mathbf{y})^{2} + \left(1 - \mathrm{SCS}(\mathbf{x},\mathbf{y})\right)^{2}} \\
\text{Spectral angle mapper:} \quad & \mathrm{SAM}(\mathbf{x},\mathbf{y}) = \arccos\!\left(\frac{\mathbf{x}\cdot\mathbf{y}}{\lVert \mathbf{x}\rVert \lVert \mathbf{y}\rVert}\right)
\end{align}

with SAM defined per \citet{kruse1993sam}. Dynamic time warping (DTW) \citep{sakoe1978dtw} and its derivative variant (DDTW) \citep{keogh2001derivative} were computed via a constrained warping-path optimisation over $\mathbf{x}$ and $\mathbf{y}$ and over their first differences $\Delta\mathbf{x}, \Delta\mathbf{y}$ respectively. Spectral information divergence (SID) was computed as a symmetric Kullback--Leibler divergence between $\mathbf{x}$ and $\mathbf{y}$ rescaled to valid probability distributions, following \citet{chang2000sid}:
\begin{equation}
\mathrm{SID}(\mathbf{x},\mathbf{y}) = \sum_{i} p_{i}\log\frac{p_{i}}{q_{i}} + \sum_{i} q_{i}\log\frac{q_{i}}{p_{i}}, \qquad p_{i} = \frac{x_{i}-\min(\mathbf{x})+\epsilon}{\sum_{j}\left(x_{j}-\min(\mathbf{x})+\epsilon\right)}
\end{equation}
with $q_{i}$ defined analogously for $\mathbf{y}$ and $\epsilon$ a small positive constant preventing degeneracy for near-constant profiles.

\begin{figure}[H]
\centering
\includegraphics[width=\textwidth, keepaspectratio]{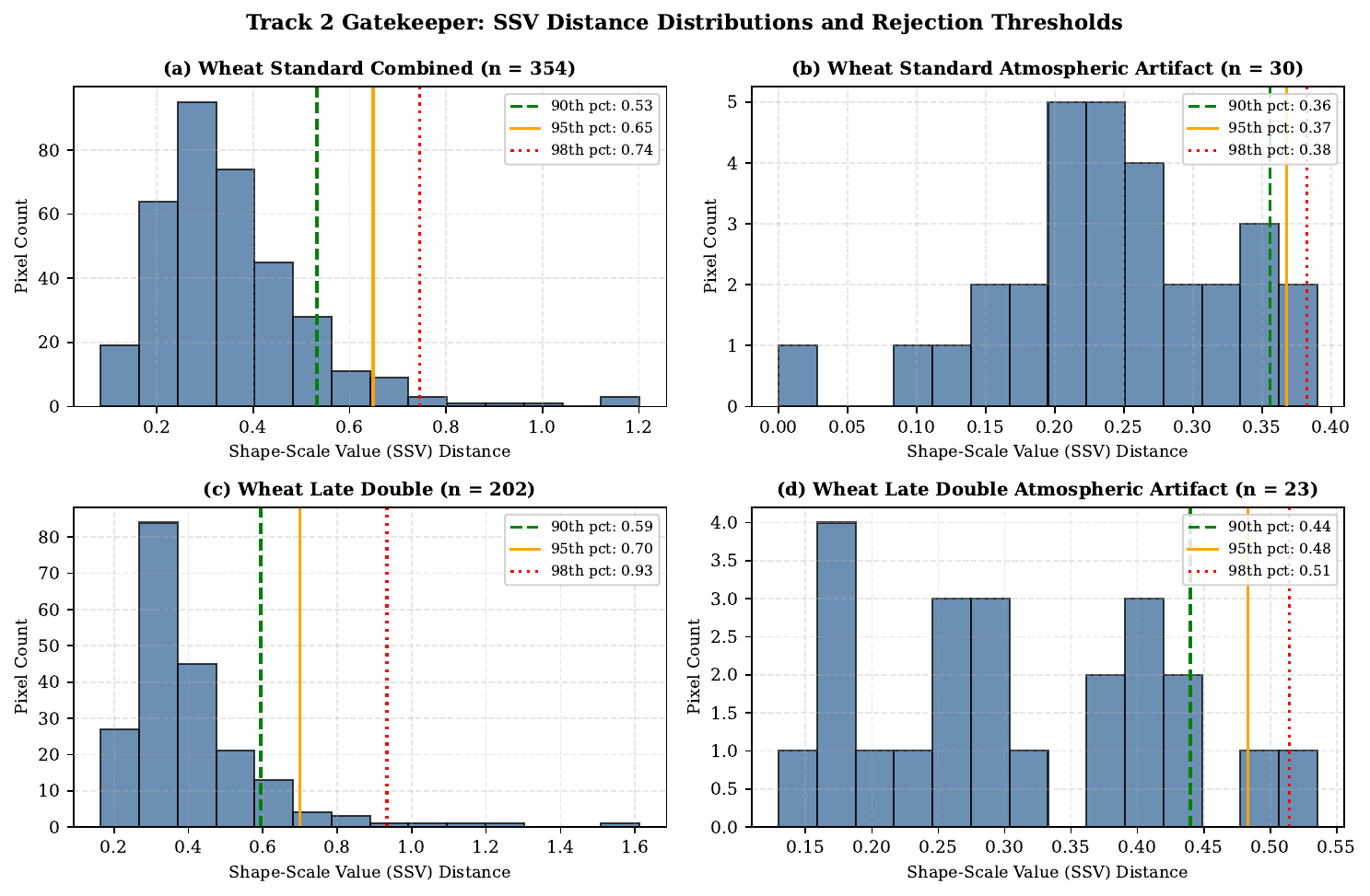}
\caption{SSV distance-score distributions and percentile lines for the four wheat-related classes, computed on the final 849-sample dataset for display. Dashed, solid, and dotted vertical lines mark the 90th, 95th, and 98th percentiles respectively; the operational gatekeeper thresholds were calibrated during rule development on earlier sample sets (Sections~\ref{sec:batch2} and~\ref{sec:batch3}), not on the data shown here.}
\label{fig:ssvdist}
\end{figure}

Of these six metrics, SSV is the sole distance used by the gatekeeper stage of the rule-based appeals mechanism (Section~\ref{sec:batch3}). SSV is preferred over either of its constituent terms in isolation because it penalises magnitude and shape divergence jointly: a candidate curve offset in overall NDVI amplitude but phenologically synchronous with its centroid (high SCS, moderate ED) and a candidate curve of matching amplitude but phenologically desynchronised (low SCS, low ED) are each individually under-penalised by ED or SCS alone, whereas SSV's Euclidean combination of the two penalises both failure modes simultaneously. This combined formulation has documented precedent for exactly this kind of curve-matching task: \citet{xu2018ssv} adopted the same SSV measure for temporal-model matching in Sentinel-1 SAR-based crop classification, reporting it as an effective discriminator of intensity-curve similarity across a multi-temporal classification scheme structurally analogous to the one employed here. Figure~\ref{fig:ssvdist} shows the empirical SSV distance-score distribution, together with the 90th, 95th, and 98th percentile thresholds derived from it, for the four wheat-related classes in the curated 849-sample dataset.

\subsubsection{Dynamic centroid library and iterative self-correction}
\begin{figure}[H]
\centering
\includegraphics[width=\textwidth, keepaspectratio]{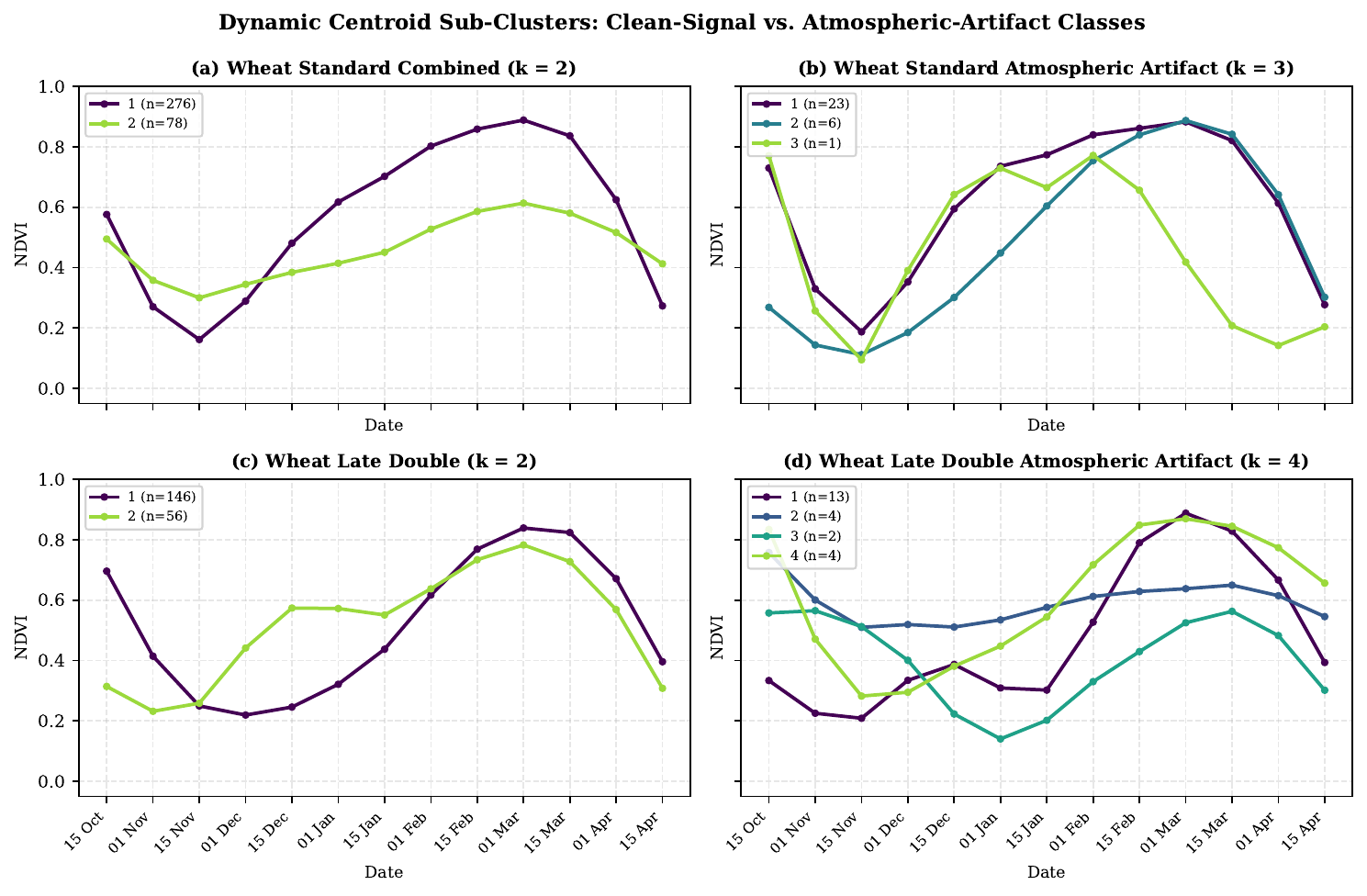}
\caption{Dynamic centroid sub-clusters computed by $k$-means clustering on the final 849-sample dataset, shown for display, for clean-signal versus atmospheric-artifact wheat classes. Atmospheric-artifact classes required a larger optimal $k$ (3--4) than their clean-signal counterparts ($k=2$), visible here as greater divergence among sub-cluster curves.}
\label{fig:centroidcomp}
\end{figure}

Sub-class temporal archetypes ("centroids") were derived by $k$-means clustering of gold-standard class-member NDVI curves, with $k$ selected per class by maximising the silhouette coefficient over a candidate range $k \in [2, \min(10, \lfloor n_{\mathrm{class}}/2 \rfloor)]$; classes with fewer than four members were represented by their arithmetic mean curve without clustering. The rule engine's initial library was built from the Batch 2 gold-standard set, and its first iteration comprised 32 centroids across the twelve working classes of that set. For display, the same procedure was re-run on the final 849-sample dataset, yielding 27 centroids across the twelve master classes (Figure~\ref{fig:fullgrid}), with atmospheric-artifact subclasses consistently requiring a larger optimal $k$ (3--4) than their corresponding clean-signal counterparts ($k=2$), reflecting the greater within-class heterogeneity introduced by residual cloud contamination. This contrast is illustrated directly in Figure~\ref{fig:centroidcomp}, which compares the sub-cluster centroid curves for both wheat sowing-window classes against their atmospheric-artifact counterparts.

These initial centroids and their associated percentile-based gatekeeper thresholds were applied back to the Batch 2 set itself as an internal consistency check, surfacing 35 unexpected rejections among samples already manually confirmed as correctly labelled. Investigation attributed these rejections to legitimate start-of-season (SOS) and end-of-season (EOS) timing shifts not accommodated by the initial threshold scheme. This finding motivated the SOS/EOS-shift pardon rule described in Section~\ref{sec:batch3} and a revised centroid library (32--37 centroids across iterations). This self-correction cycle is disclosed explicitly as an iterative tuning step performed against the Batch 2 set; Section~\ref{sec:limitations} discusses its implications for the independence of rule-engine performance estimates.

\subsection{Batch 3: Scaled Application and Class Simplification}
\label{sec:batch3}

The revised centroid library and rule engine were applied at scale to Batch 3 (1{,}000 candidate points). Points falling outside the Punjab State boundary were removed, yielding 980 points; geographic verification against a 25\,km buffered state boundary then excluded a further 198 points (183 beyond the buffer and 15 anomalous samples outside Punjab), retaining 782.

\subsubsection{Rule-based appeals mechanism}
Candidate samples whose best-matching centroid distance exceeded a per-centroid gatekeeper threshold were passed to a three-stage appeals procedure prior to rejection: a cloud-overlap pardon, admitting samples with high temporal-envelope overlap against the matched centroid as an atmospheric-artifact variant of that class; a spectral-angle pardon, admitting low-SAM, low-amplitude samples as a sparse-signal variant; and the start-of-season (SOS) shift pardon motivated in Section~\ref{sec:batch2}, admitting samples whose season-onset timestep (first index at which NDVI $\geq 0.35$) differed from the matched centroid's onset by one to five timesteps ($\approx$2--10 weeks), reflecting biologically plausible sowing-date variability rather than misclassification. A subsequent agronomic override stage reassigned wheat-matched candidates exhibiting a low seasonal peak with early decline to a potato/short-Rabi label, and candidates exhibiting an abrupt mid-February-to-March NDVI collapse to a mustard label, consistent with each crop's known senescence timing in the study region.

Of the 782 geo-verified Batch 3 candidates, 56 (7.2\%) were individually re-verified through visual comparison of each pixel's NDVI profile against corresponding Google Earth Engine satellite imagery, with the remaining 726 (92.8\%) retained under the rule engine's automated assignment. The working taxonomy of the 782 samples comprised 44 labels, including atmospheric-artifact, sparse-signal, and early- or late-sown sub-variants. For downstream model training these were mapped to eleven master classes, and samples whose working label fell outside the mapping were dropped, leaving 717 of the 782 samples. The 65 dropped samples comprised 30 wheat sub-variants (18 carrying an intensive rice--wheat label and 12 late-sown, early-sown, or sparse-signal variants outside the master mapping) and 35 samples of minor crop and land-cover labels, none with more than four samples per label (for example sugarcane, cotton/fallow, fallow, vegetables/potato, and rabi pulses). Eleven of the twelve final master classes are represented in the 717-sample Batch 3 output; the twelfth, mustard, is not, because the single mustard-labelled Batch 3 sample fell outside the mapping, and it enters the final training set through the hard-case extraction described in Section~\ref{sec:hardcases}.

\subsection{Excel Ground-Truth Integration and Hard-Case Extraction}
\label{sec:hardcases}

An independently sourced, point-level ground-reference dataset -- recording crop name, sowing date, and growth-stage information alongside geographic coordinates across multiple Rabi seasons -- was obtained as an additional, higher-detail validation resource. Records were filtered to the 2021-22 Rabi season and to Rabi-season crop types, yielding 874 candidate points. Two candidate rule-engine configurations were evaluated against this set: the Batch 3 threshold dictionary, which rejected the substantial majority of points even after relaxation, and the Batch 2 threshold dictionary paired with the revised centroid library, which was retained on the basis of its markedly better agreement with this independent reference (477 perfect matches, 363 admitted via the appeals mechanism, 34 rejected, of 874 total).

To identify hard cases, the rule engine (Batch 3 threshold dictionary and centroid library) was applied to a 575-point subset of the 874 surveyor points (479 wheat, 86 potato, 8 mustard, and 2 other-crop points), and its assigned classes were compared with the surveyor crop labels. Three extraction rules were applied: all mustard-labelled points except one sample identified as exhibiting severe spectral ambiguity (the rule engine labelled none of the eight mustard points as mustard); potato-labelled points to which the rule engine assigned one of seven labels (fallow/weeds, short-Rabi/zaid, unclassified anomaly, or a sparse-signal or early-sown wheat variant); and wheat-labelled points to which the rule engine assigned a fallow or fodder label. This extraction yielded 132 hard-case samples: 7 mustard-labelled points, 47 potato and short-Rabi-labelled points, and 78 wheat (standard, combined)-labelled points. They were retained because the rule engine's automated label for them was unreliable or disagreed with the surveyor record, and in the final training set they carry the surveyor's crop label rather than the rule-engine label. The 575-point subset was also used while relaxing and tuning the rule engine, so it is not independent of that engine. The mustard component (7 samples) constitutes the entirety of the mustard class in the final training set, since the single mustard-labelled Batch 3 sample was excluded by the eleven-class mapping (Section~\ref{sec:batch3}).

\subsection{Final Training Set Assembly}
\label{sec:finalassembly}

The final 849-sample training set was assembled by merging the 717-sample Batch 3 output (Section~\ref{sec:batch3}) with the 132-sample hard-case extraction (Section~\ref{sec:hardcases}). Batch 2's approximately 324 manually verified gold-standard samples (Section~\ref{sec:batch2}) are not themselves present as rows in this final training set; their role in the pipeline is confined to constructing the centroid library and threshold dictionary used to process Batches 2 and 3, a design choice discussed further as a disclosed limitation in Section~\ref{sec:limitations}.

\subsection{Track 1: Operational XGBoost Classifier}
\label{sec:xgboost}

The operational classifier consumes the thirteen-dimensional Savitzky--Golay-smoothed NDVI vector as its sole feature input, without ancillary SAR or auxiliary bands. The curated 849-sample dataset was partitioned into training and validation subsets via stratified random sampling (test proportion $=0.20$, fixed random seed), yielding 679 training and 170 validation samples across the twelve master classes described in Section~\ref{sec:studyarea}. To address class imbalance, the Synthetic Minority Over-sampling Technique \citep{chawla2002smote} was applied exclusively to the training partition, with the neighbourhood parameter dynamically bounded by the smallest training-class cardinality to avoid degenerate interpolation; the validation partition remained untouched by resampling throughout. The validation partition was, however, also passed to the early-stopping routine of the XGBoost classifier and of the neural baselines, so the Tier~1 accuracies are mildly optimistic relative to a fully untouched test set.

The classifier is a multi-class Extreme Gradient Boosting ensemble \citep{chen2016xgboost}, configured with 500 estimators, a learning rate of 0.05, a maximum tree depth of 6, row and column subsampling ratios of 0.8, the histogram-based tree construction method, an early-stopping patience of 20 rounds, and the multinomial softmax-probability objective. Following training, the fitted model was deployed pixel-wise across the full study district raster (36{,}255{,}347 valid pixels), with non-agricultural urban and water areas subsequently masked using standard boundary layers, to generate the district-wide crop classification product evaluated in Section~\ref{sec:results-forward}.

\subsection{Spatially-Blocked Cross-Validation}
\label{sec:spcv}

The stratified random 679/170 split underlying the Tier 1 validation accuracy (Section~\ref{sec:xgboost}) does not enforce geographic separation between training and validation samples; as disclosed in Section~\ref{sec:studyarea}, spatially proximal samples allocated to opposite sides of this split may share correlated phenological noise, risking a modest optimistic bias in the reported accuracy. To test this directly, a spatially-blocked cross-validation (SpCV) was conducted as a robustness check on the full 849-sample curated dataset, using the twelve-class taxonomy established in Section~\ref{sec:studyarea}.

Sample coordinates were clustered into five spatial blocks via $k$-means clustering directly on coordinate pairs, and five-fold cross-validation was performed using \texttt{GroupKFold}, constraining each spatial block to appear entirely within either the training or the held-out fold for a given iteration, eliminating any possibility of geographic overlap between train and test partitions. The XGBoost configuration matched the production configuration of Section~\ref{sec:xgboost} (learning rate 0.05, maximum depth 6, row and column subsampling ratios of 0.8, histogram-based tree construction) except that the number of estimators was fixed at 200 and early stopping was not used, so no held-out fold influenced model selection, and SMOTE was applied strictly within each training fold, as in the original pipeline. Because spatial clustering operates on coordinates independently of class label, and because clustering into five blocks distributed even the rarest classes (Mustard, $n=7$; Forest/Tree Cover, $n=6$) such that at least one representative of every class remained present in every training fold across all five iterations, no fold required any form of synthetic or placeholder training data to satisfy the classifier's class-indexing requirement.

\section{Results}
\label{sec:results-forward}

\subsection{Feature Sensitivity Validation}
\label{sec:featuresens}

Table~\ref{tab:featuresens} reports, for all thirteen NDVI timesteps, the standardised Cohen's $d$ effect size for wheat versus non-wheat separability computed directly on the curated 849-sample dataset ($n_{\mathrm{wheat}}=609$, $n_{\mathrm{non\text{-}wheat}}=240$), alongside the XGBoost gain-based feature importance from the fitted Track 1 classifier. Ranks are reported independently for each metric, ordered by descending absolute magnitude.

\begin{table}[htbp]
\centering
\caption{Feature sensitivity: Cohen's $d$ (wheat vs.\ non-wheat, $n=849$) versus XGBoost gain-based importance, all thirteen NDVI timesteps.}
\label{tab:featuresens}
\begin{tabular}{llrrrr}
\toprule
Timestep & Date & Cohen's $d$ & $|d|$ rank & XGBoost gain & Gain rank \\
\midrule
$t_{10}$ & 01 Mar & $2.214$  & 1  & 0.1028 & 3 \\
$t_{11}$ & 15 Mar & $2.049$  & 2  & 0.0705 & 8 \\
$t_{9}$  & 15 Feb & $1.853$  & 3  & 0.0709 & 7 \\
$t_{8}$  & 01 Feb & $1.347$  & 4  & 0.1195 & 2 \\
$t_{12}$ & 01 Apr & $1.126$  & 5  & 0.0948 & 4 \\
$t_{3}$  & 15 Nov & $-1.088$ & 6  & 0.0618 & 10 \\
$t_{4}$  & 01 Dec & $-0.817$ & 7  & 0.0731 & 6 \\
$t_{7}$  & 15 Jan & $0.758$  & 8  & 0.0699 & 9 \\
$t_{13}$ & 15 Apr & $-0.599$ & 9  & 0.0383 & 12 \\
$t_{1}$  & 15 Oct & $0.513$  & 10 & 0.0367 & 13 \\
$t_{2}$  & 01 Nov & $-0.507$ & 11 & 0.0785 & 5 \\
$t_{6}$  & 01 Jan & $0.313$  & 12 & 0.1219 & 1 \\
$t_{5}$  & 15 Dec & $-0.184$ & 13 & 0.0614 & 11 \\
\bottomrule
\end{tabular}
\end{table}

The two metrics agree that the composite February--April window carries the majority of discriminative signal: four of the five highest-ranked timesteps under Cohen's $d$ ($t_{8}$--$t_{12}$) also fall within the top six under XGBoost gain. They diverge markedly, however, on the single most important timestep and on the least important: Cohen's $d$ identifies 01 March ($t_{10}$) as the point of maximum wheat/non-wheat separability, reflecting the grain-fill phenological stage at which wheat's NDVI trajectory diverges most sharply from the heterogeneous non-wheat aggregate; XGBoost gain instead assigns its single highest weight to 01 January ($t_{6}$), a timestep Cohen's $d$ ranks twelfth of thirteen. This divergence is reported rather than reconciled: the two metrics answer different questions -- Cohen's $d$ measures univariate class separability in isolation, while gain-based importance reflects a feature's marginal contribution to a nonlinear ensemble's split decisions conditional on all other features -- and their disagreement is itself informative evidence that no single timestep, and no single importance metric, should be treated as uniquely authoritative.

\subsection{Baseline Model Comparison}
\label{sec:baselinecomp}

All four classifiers were evaluated under an identical stratified 679/170 train/validation partition of the 849-sample curated dataset. Table~\ref{tab:baselines} reports overall accuracy alongside each architecture's trainable parameter count, where applicable.

\begin{table}[htbp]
\centering
\caption{Classifier comparison under identical 679/170 stratified partition ($n=170$ validation samples).}
\label{tab:baselines}
\begin{tabular}{lrr}
\toprule
Model & Trainable parameters & Overall accuracy \\
\midrule
1D-CNN & 9{,}932 & 66.47\% \\
LSTM & 30{,}764 & 64.12\% \\
Hybrid CNN--LSTM & 14{,}124 & 64.71\% \\
XGBoost & --\textsuperscript{*} & \textbf{78.82\%} \\
\bottomrule
\multicolumn{3}{l}{\rule{0pt}{3ex}\footnotesize \textsuperscript{*}Tree-based ensemble (up to 500 trees, early-stopped; max. depth 6).} \\
\end{tabular}
\end{table}

XGBoost outperforms all three deep-learning baselines by a margin of at least 12 percentage points despite -- and, we argue, in part because of -- possessing no directly comparable dense parameter matrix. The LSTM baseline is notable for pairing the largest parameter count of the four architectures (30{,}764) with the lowest accuracy, an outcome consistent with two compounding constraints specific to this task: a short input sequence ($T=13$ timesteps), which affords recurrent architectures little opportunity to exploit long-range temporal dependencies relative to the sequence lengths for which they are typically designed, and a small training partition ($n=679$), under which a 30{,}764-parameter recurrent network is comparatively exposed to overfitting relative to a shallow, axis-aligned tree ensemble operating directly on thirteen discrete, phenologically interpretable features.
Dropout (rates 0.2--0.3) and early stopping on validation loss with best-weight restoration were applied to the 1D-CNN, LSTM, and hybrid CNN--LSTM baselines; no weight decay, learning-rate scheduling, or hyperparameter search specific to this study's sample size was performed. This is disclosed explicitly because the resulting comparison in Table~\ref{tab:baselines} should be read as a comparison under representative default conditions for each architecture family, rather than as an exhaustively hyperparameter-optimised comparison across all four models.

\subsection{The Two-Tier Validation Cascade}
\label{sec:threetier}

The operational classifier was evaluated at three progressively broader scopes, summarised in Table~\ref{tab:threetier}. Tier 1 reports the accuracy of the production classifier on the 170-sample validation partition (Section~\ref{sec:xgboost}), which was also used for early stopping. Tiers 2A and 2B report deployment-scale performance across, respectively, a balanced stratified sample of the district raster and the full valid-pixel census.

\begin{table}[htbp]
\centering
\caption{Two-tier validation cascade. Em-dashes denote metrics not computed at that tier.}
\label{tab:threetier}
\resizebox{\textwidth}{!}{%
\begin{tabular}{llrrrrrr}
\toprule
Tier & Scope & $n$ & OA & Precision & Recall & F1 & Kappa \\
\midrule
1 & Validation partition & 170 samples & 78.82\% & --- & --- & --- & --- \\
\midrule
2A & Balanced stratified sample & 500{,}000 px & 62.12\% & 60.26\% & 71.15\% & --- & 0.2424 \\
2B & Full population census & 36{,}255{,}347 px & 67.56\% & 86.31\% & 71.05\% & 0.7794 & 0.1882 \\
\bottomrule
\end{tabular}%
}
\end{table}
\noindent The Tier 2B configuration additionally yields an Intersection-over-Union of 0.6385. Overall accuracy is 78.82\% at Tier 1, 62.12\% at Tier 2A, and 67.56\% at Tier 2B. Tier 1 is a twelve-class accuracy on curated samples, whereas Tiers 2A and 2B are binary comparisons against the spatial mask, so the two are not directly comparable. Tier 2A drew 250{,}000 pixels from each reference class, so its overall accuracy is the mean of recall and specificity and its precision depends on that class-balanced design. Recall is 71.15\% at Tier 2A and 71.05\% at Tier 2B, and specificity is about 53\% in both (derived for Tier 2B from the reported overall accuracy, recall, and mask prevalence), so the difference in overall accuracy between the two tiers reflects class weighting, not a change in classifier behaviour. The lower agreement of the deployment tiers relative to Tier 1 may in part reflect a validation-to-deployment domain shift: a plausible contributor is that the curated 849-sample set is, by design, enriched for phenologically distinctive and previously ambiguous cases, whereas the full district raster additionally contains substantial areas of spectrally intermediate or mixed-pixel land cover absent from the curated set. Tier 2B's precision and recall are computed on the full population and are therefore the operationally definitive figures reported in Section~\ref{sec:acreage}.

It is important to state explicitly what Tier 2B's precision and recall do, and do not, establish. Both are computed relative to the government spatial reference mask -- the same product shown in Section~\ref{sec:acreage} to diverge from the official tabular target by $+25.11\%$. Because that mask is itself demonstrated to carry a systematic positive area bias, Tier 2B's precision and recall should not be read as an independent, bias-free validation of pixel-level classification accuracy; a false positive or false negative computed against an over-inclusive reference does not necessarily indicate a genuine classification error. Tier 2B is reported here as a population-scale deployment diagnostic consistent with prior operational practice, not as evidence that resolves questions of pixel-level accuracy independent of the spatial mask's own documented bias. The spatially-blocked cross-validation reported in Section~\ref{sec:spcvresults}, evaluated against ground-truth labels rather than the spatial mask, is the more appropriate check for that purpose.

\subsection{Spatial Robustness Check}
\label{sec:spcvresults}

Table~\ref{tab:spcv} reports per-fold accuracy under the spatially-blocked cross-validation scheme described in Section~\ref{sec:spcv}.

\begin{table}[htbp]
\centering
\caption{Spatially-blocked five-fold cross-validation accuracy, full 849-sample dataset (twelve-class taxonomy), geographically disjoint train/test folds via $k$-means spatial clustering and \texttt{GroupKFold}. The mean is the unweighted mean of the five fold accuracies, and $\pm$ is the population standard deviation across folds ($\mathrm{ddof}=0$). The pooled accuracy over all 849 held-out samples is 75.62\%.}
\label{tab:spcv}
\begin{tabular}{lrr}
\toprule
Fold & $n$ (held-out block) & Accuracy \\
\midrule
1 & 222 & 77.93\% \\
2 & 164 & 74.39\% \\
3 & 163 & 76.69\% \\
4 & 156 & 73.72\% \\
5 & 144 & 74.31\% \\
\midrule
\textbf{Mean $\pm$ SD} & 849 & \textbf{75.41\% $\pm$ 1.62\%} \\
\bottomrule
\end{tabular}
\end{table}

The spatially-blocked mean accuracy (75.41\%) is $3.41$ percentage points lower than the Tier 1 stratified random-split accuracy (78.82\%, Section~\ref{sec:baselinecomp}). This direction and magnitude of decline is consistent with the presence of a modest optimistic bias in the original split attributable to spatial autocorrelation, as anticipated in Section~\ref{sec:limitations}: eliminating geographic overlap between training and validation samples removes the classifier's ability to exploit within-field or near-field phenological similarity across the train/validation boundary, and the resulting accuracy should be read as the more conservative, geographically independent estimate of the classifier's discriminative performance on unseen locations within Punjab State. Every training fold retained at least one representative of all twelve classes, including the rarest (Mustard, Forest/Tree Cover), so the twelve-class task matches that of Section~\ref{sec:baselinecomp}, although the tree budget differs (fixed here, early-stopped for the production model). The relatively small and consistent magnitude of the decline across all five geographically disjoint folds (population standard deviation 1.62\%) indicates that the classifier's performance is not predominantly an artefact of spatial leakage, though this check remains confined to spatial independence among the Punjab reference samples and does not by itself establish generalisation to other districts, seasons, or agro-climatic zones, which is addressed separately in the external-validity limitation (Section~\ref{sec:limitations}).

\subsection{Metric Transparency and the Kappa Paradox}
\label{sec:kappaparadox}

The Tier 2B Kappa coefficient of 0.1882 falls within the "slight" agreement band on the conventional Landis--Koch interpretive scale, a figure that -- read in isolation and without reference to Tier 2B's substantially higher precision (86.31\%) and F1-score (0.7794) -- risks being misread as evidence of weak classifier performance. It is not; the apparent tension is a well-documented mathematical property of Cohen's Kappa under skewed class marginals, occasionally termed the kappa paradox \citep{feinstein1990kappa}, and it is addressed explicitly here rather than left for a reader to discover unexplained.

Cohen's Kappa is defined as $\kappa = (p_{o} - p_{e})/(1 - p_{e})$, where $p_{o}$ is observed agreement (here, overall accuracy) and $p_{e}$ is the agreement expected under independence given the classifier's and the reference data's marginal class proportions. Rearranging the Tier 2B figures ($p_{o}=0.6756$, $\kappa=0.1882$) yields an implied chance-agreement rate of $p_{e}\approx0.60$. This elevated $p_{e}$ is a direct consequence of class prevalence at deployment scale: wheat is the dominant land cover across the study district in both the classifier's output and the administrative reference, comprising approximately 66.4\% of valid district pixels under the XGBoost prediction (240{,}682.51\,ha of 362{,}553.47\,ha total) and 80.65\% in the spatial reference mask against which Tier 2B is computed (292{,}386.91\,ha). When both the predicted and true marginals concentrate this heavily on a single class, the chance-agreement term $p_{e}$ -- which scales with the sum of the squared marginal proportions -- rises sharply, mechanically compressing the achievable Kappa ceiling regardless of underlying classification quality. Precision and recall, computed conditional on the predicted- and true-positive subsets respectively, are not subject to this compression and are the more diagnostic metrics for this task's operational objective: area estimation for a known majority class, rather than balanced multi-class discrimination.

This prevalence-driven explanation should not, however, be read as the sole driver of the depressed Kappa coefficient. The Tier 2B confusion matrix also indicates genuine misclassification within the heterogeneous non-wheat land-cover aggregate itself. Because the Tier 2B reference product is binary (wheat versus non-wheat), this study cannot decompose that residual confusion into specific class pairs at deployment scale; at the curated 849-sample level, however, spectrally and phenologically similar minor Rabi classes -- mustard and potato/short-Rabi being plausible candidates given their comparatively small representation in the curated training set (Section~\ref{sec:studyarea}) -- are natural candidates for this kind of within-non-wheat confusion, though confirming this specific attribution would require a full multi-class confusion matrix at population scale, which lies beyond the binary reference product available for this study. The low Tier 2B Kappa is therefore best read as the joint product of two effects: the mechanical prevalence-driven compression derived above, and a residual, currently unresolved component of genuine non-wheat misclassification.

\subsection{District Acreage Reconciliation}
\label{sec:acreage}

Table~\ref{tab:acreage} reconciles the Tier 2B classification product against the official government tabular target and against the government's own spatial reference mask for the same season and district.

\begin{table}[htbp]
\centering
\caption{District wheat-area reconciliation, 2022 Rabi season, Patiala district (total district area 362{,}553.47\,ha).}
\label{tab:acreage}
\resizebox{\textwidth}{!}{%
\begin{tabular}{lrrr}
\toprule
Source & Area (ha) & \% of district & Deviation vs.\ tabular target \\
\midrule
Official tabular target & 233{,}700.00 & 64.46\% & --- (reference) \\
XGBoost (Track 1) prediction & 240{,}682.51 & 66.39\% & $+2.99\%$ \\
Government spatial mask (raw) & 292{,}386.91 & 80.65\% & $+25.11\%$ \\
\bottomrule
\end{tabular}%
}
\end{table}
The operational classifier's prediction deviates from the official tabular target by $+2.99\%$, a margin consistent with routine survey and boundary-delineation variance. The government's own spatial reference mask, by contrast, deviates from the identical tabular target by $+25.11\%$ -- a substantial \emph{positive area bias} (spatial overestimation) that is independent of, and unrelated to, the classifier evaluated in this study. This asymmetry is the main result of the study: a classifier trained on a small, auditor-curated, phenologically validated sample set reconciles more closely with the administrative record of wheat area than does the spatial product conventionally treated as ground truth for validating such classifiers.

This agreement concerns net district acreage only. Relative to the spatial mask, the reported precision (86.31\%) and recall (71.05\%) imply approximately 84{,}600\,ha of mask-labelled wheat not predicted and approximately 32{,}900\,ha predicted where the mask records none, a net difference of approximately $-51{,}700$\,ha. Because the mask is itself biased and no independent probability sample of field-verified points is available at district scale, this study cannot determine how much of this pixel-level disagreement reflects classifier error and how much reflects mask error; the $+2.99\%$ figure should not be read as evidence of pixel-level accuracy.

\section{Discussion}
\label{sec:discussion}

Beyond the results reported in Section~\ref{sec:results-forward}, one interpretive question raised by the district acreage reconciliation (Section~\ref{sec:acreage}) merits discussion. We evaluate both the classifier and the spatial mask against the official tabular target (233{,}700.00\,ha), treating it as the authoritative baseline. We acknowledge that administrative tabular data is itself subject to institutional reporting biases and should not be confused with flawless absolute truth; however, because it remains the official metric by which district crop production is formally quantified, this study still evaluates against it as the reconciliation baseline. One hypothesis for the magnitude of the spatial mask's $+25.11\%$ overestimation concerns how such records are compiled. Administrative wheat-area records are commonly compiled from manual field enumeration (the \emph{girdawari} system); whether the mask evaluated here derives from such enumeration was not established, so the following is a hypothesis. If field boundaries were recorded by eye or on foot, they may have been generalised outward, incorporating adjoining roads, field margins, and small barren patches. Such an aggregation effect would produce a one-directional positive area bias of the kind observed here, unlike pixel-level assignment by a satellite-derived classifier. This study contains no test of the hypothesis. The remainder of this section turns from this interpretive question to three further empirical observations arising directly from the results reported in Section~\ref{sec:results-forward}.

First, per-class precision for the deep-learning baselines is not uniformly below that of XGBoost. For the forest (combined) class ($n=14$ validation samples), the 1D-CNN reached a precision of 1.00 against 0.92 for XGBoost, with lower recall (0.71 against 0.79). At these support sizes the difference amounts to one or two samples and is not interpreted further.

Second, the XGBoost gain-based feature importance reported in Section~\ref{sec:featuresens} peaks sharply across the January--March window (01 January, 01 February, and 01 March collectively account for the three highest gain scores) while assigning its two lowest weights to the sequence's edges, 15 October and 15 April. This is broadly, though not exactly, consistent with the independent Cohen's $d$ ranking, and together the two metrics support an agronomic reading in which the mid-to-late-season vegetative and grain-fill window carries the operationally decisive classification signal, while the earliest post-sowing and latest pre-harvest observations -- where wheat's spectral signature is least differentiated from bare soil or senescent ground cover, respectively -- contribute comparatively little standalone discriminative value.

Third, the one-to-five-timestep tolerance window built into the Track 2 start-of-season pardon (Section~\ref{sec:batch3}), corresponding to approximately two to ten weeks at the study's 15-day compositing interval, is not an arbitrary tuning choice but an explicit, testable encoding of the sowing-date variability documented across smallholder Rabi wheat cultivation in the study region, empirically motivated by the Batch 2 self-correction cycle described in Section~\ref{sec:batch2}.

Taken together, these findings support the dual-track design principle motivating this study: curation-stage quality control (Track 2) and deployment-stage classification (Track 1) benefit from being treated as distinct problems with distinct evaluation criteria, rather than as a single undifferentiated pipeline whose accuracy is reported at only one scale.

\subsection{Limitations and Future Work}
\label{sec:limitations}

The framework presented in this study carries eight specific, data-backed limitations that constrain the strength of the claims that can be drawn from it and motivate concrete directions for subsequent work.

\textbf{Sampling design and spatial autocorrelation.} As disclosed in Section~\ref{sec:studyarea}, the sampling procedure underlying the 849-sample curated dataset enforced no minimum inter-point spacing constraint at the initial seeding stage, raising the possibility that the Tier 1 stratified random-split accuracy (78.82\%) carries a modest optimistic bias from spatially proximal train/validation samples. This was tested directly via a spatially-blocked cross-validation (Section~\ref{sec:spcvresults}), which found a spatially independent accuracy of 75.41\% (fold mean; $\pm$ 1.62\%, population standard deviation across folds) across the full twelve-class taxonomy -- a 3.41 percentage point decline consistent with a modest such bias, with the classifier's performance otherwise holding stable across all five geographically disjoint folds. The Tier 1 figure should accordingly be read as a mild upper estimate rather than a spatially independent one; the spatially-blocked figure is the more conservative estimate of held-out performance within Punjab State.

\textbf{Statistical fragility of minority-class metrics.} The water class carries a validation support of only $n=2$ samples; the 1.00 precision of XGBoost on this class reflects zero false positives across a single true positive (recall $=0.50$) and is not a statistically stable estimate of class-conditional reliability. This is not an isolated case: of the twelve classes represented in the 170-sample validation fold, mustard and forest/tree cover each carry a support of exactly one sample and water a support of two, with several further classes (barren, both atmospheric-artifact wheat subclasses) supported by fewer than seven samples. Precision, recall, and F1 figures for these minority classes throughout this study should be read as indicative rather than statistically stable, and any deployment decision keyed to minority-class performance should be validated against a materially larger held-out sample before being relied upon operationally.

\textbf{Iterative rule-tuning against an overlapping evaluation pool.} As disclosed in Section~\ref{sec:batch2}, the rule engine's gatekeeper thresholds and pardon logic were revised in direct response to anomalies discovered by applying an early version of the engine back to the Batch 2 set that had itself calibrated that engine. This iterative self-correction produced a materially improved rule engine, but any accuracy or agreement figure attributed to the rule engine's performance on Batch 2 specifically should be read with the understanding that the engine was tuned against that same set, rather than treated as an independent evaluation of rule-engine performance.

\textbf{Exclusion of the Batch 2 gold-standard set from final training.} As disclosed in Section~\ref{sec:finalassembly}, the approximately 324 manually verified Batch 2 samples contributed to centroid and threshold construction but are not present as rows in the final 849-sample training set used to fit the Track 1 classifier. This represents unused, high-confidence labelled data; incorporating it directly into a future training set is a natural extension of the current framework.

\textbf{Rigidity of the rule-based auditor.} The Track 2 appeals mechanism (Section~\ref{sec:batch3}) resolves phenological edge cases through hard, per-centroid distance thresholds and a fixed sequence of pardon rules. This design is transparent and auditable, but brittle at the threshold boundary: a candidate sample falling marginally outside a gatekeeper threshold is rejected identically to one falling far outside it, and the relative weight assigned to each of the six distance metrics is fixed rather than learned from data. A probabilistic weak-supervision reformulation -- in which each distance metric and pardon rule is instead expressed as a labelling function within a data-programming framework, with a generative label model learning per-function reliability weights and producing calibrated soft labels rather than hard accept/reject decisions -- would allow the curation stage to retain the same six metrics currently in use while degrading gracefully at class boundaries, rather than discarding borderline-informative samples outright.

\textbf{Unexploited multi-modal signal.} Sentinel-1 SAR backscatter is not used in the reported framework: the Track 1 classifier is trained exclusively on the thirteen-dimensional optical NDVI vector. This design choice is a likely contributor to the atmospheric-artifact subclasses documented in Section~\ref{sec:batch2}: because the classifier's sole input is cloud-sensitive optical reflectance, residual cloud contamination in a minority of acquisitions produces NDVI curves anomalous enough to require dedicated wheat-standard-atmospheric-artifact and wheat-late/double-atmospheric-artifact labels rather than being resolved directly. Extending the Track 1 feature space to include co-registered VV and VH backscatter -- an all-weather signal unaffected by cloud cover -- alongside the existing NDVI vector is a direct route to eliminating the need for these artifact-specific subclasses altogether, collapsing the twelve-class taxonomy toward a more parsimonious ten-class structure and removing a source of label-space complexity introduced solely to compensate for a single sensor's weather sensitivity.

\textbf{Approximate distance-metric computation.} The dynamic time warping and derivative dynamic time warping distances underlying the Track 2 centroid library (Section~\ref{sec:batch2}) are computed via the \texttt{fastdtw} approximate algorithm, whose principal advantage -- near-linear-time complexity relative to the $O(n^{2})$ cost of exact DTW -- is unnecessary at the thirteen-timestep sequence length used throughout this study, for which exact DTW is computationally trivial. \texttt{fastdtw}'s approximation is documented in the time-series literature to occasionally yield warping paths inferior to the exact solution at short sequence lengths, meaning the DTW- and DDTW-based curation decisions reported in this study may not exactly reproduce those of an exact-DTW implementation. Substituting exact DTW computation -- a negligible cost at $n=13$ -- is recommended as a low-effort correction in future iterations of the framework.

\textbf{External validity.} The reference samples were drawn from across Punjab State, whereas classifier deployment and area reconciliation were carried out only within Patiala district during a single growing season, the 2022 Rabi cycle. The reference samples were not withheld by district, so the Patiala deployment is not a held-out-district evaluation, and reconciliation against tabular statistics was performed for one district only. Neither the centroid library, the rule-based appeals thresholds, nor the operational classifier's hyperparameters have been evaluated against a different agro-climatic zone or an anomalous weather year (for example, a season with atypical sowing delays, unseasonal rainfall, or heat stress affecting the wheat phenological curve). The magnitude and even the direction of the administrative reference discordance documented in Section~\ref{sec:acreage} -- and the specific numerical thresholds governing Track 2's curation logic -- should accordingly be treated as demonstrated for this one district-season combination rather than as generalisable constants. Cross-site and cross-year replication remains a necessary direction for future validation before this framework is treated as a general-purpose operational tool. Two concrete algorithmic extensions are worth testing alongside such replication. First, a bidirectional LSTM, which has demonstrated accuracies exceeding 94\% in smallholder crop classification by capturing phenological context from both temporal directions \citep{khan2024bilstm}, is a candidate replacement for the unidirectional LSTM baseline evaluated in Section~\ref{sec:baselinecomp} -- though whether its bidirectional context benefit outweighs its approximately doubled recurrent parameter count, under the same small-sample overfitting risk this study identifies for the unidirectional case, is an open empirical question rather than an assumed improvement. Second, transitioning from pixel-based to object-based classification -- grouping contiguous homogeneous pixels into field-sized objects (approximately 1--2.5\,ha, or 100--160 Sentinel-2 pixels) via time-weighted dynamic time warping prior to classification -- offers a direct, testable route to examining the boundary-generalisation hypothesis raised in Section~\ref{sec:discussion}, since object boundaries derived this way would approximate true field morphology rather than either a pixel grid or a manually walked survey boundary.

\section{Conclusion}
\label{sec:conclusion}

This study set out to quantify a specific instance of administrative reference discordance -- the divergence between administrative tabular crop statistics and the spatial reference products conventionally used to validate remotely sensed classification -- for wheat area estimation in Patiala district during the 2022 Rabi season. The proposed dual-track architecture separates this problem into an upstream diagnostic curation stage (Track 2), realised as an iterative bootstrapping procedure across three successive sample batches combining a silhouette-optimised dynamic centroid library, six complementary distance and similarity metrics, and a self-correcting rule-based appeals mechanism to assemble an 849-sample auditor-curated reference set, and a downstream operational classification stage (Track 1), an XGBoost ensemble trained exclusively on that curated set and deployed across the full 36{,}255{,}347-pixel study district. Class-balancing via SMOTE was confined strictly to the training partition at every stage of the pipeline.

The central empirical finding is that this small-sample, curation-first classifier reconciles with the official tabular wheat-area statistic to within $+2.99\%$, while the government's own spatial reference mask -- the product conventionally treated as the higher-fidelity reference -- diverges from the identical statistic by $+25.11\%$. This result should not be read as a general claim that deep-learning architectures are unsuitable for crop classification, nor that spatial reference products are inherently unreliable; rather, it demonstrates, for one district, one season, and one label-noise regime, that careful iterative sample curation can offset a substantial sample-size disadvantage relative to both larger noisy-label spatial products and higher-capacity sequence models. Whether this trade-off generalises to other districts, seasons, and crop-label taxonomies is an open empirical question, and we present the dual-track framework -- together with its documented limitations, including the disclosed spatial-autocorrelation caveat in the sampling design (Section~\ref{sec:studyarea}) and the divergent feature-importance rankings reported in Section~\ref{sec:featuresens} -- as a methodological reference for crop-area reconciliation in comparable label-noisy administrative settings, rather than as a definitive or final architecture.

\section*{CRediT Authorship Contribution Statement}
\textbf{Kasimali Agharia:} Methodology, Software, Formal analysis, Data curation, Writing - original draft, Visualization. \\
\textbf{Ujjwal Kumar Gupta:} Conceptualization, Methodology, Supervision, Validation, Resources, Project administration, Writing - review \& editing.

\section*{Acknowledgements}
The authors thank the Space Applications Centre (SAC), Indian Space Research Organisation (ISRO), for institutional support and provisioning of the foundational ground-truth data used in this study.

\section*{Data Availability Statement}
The Sentinel-2 satellite imagery supporting this research is openly available through the Google Earth Engine (GEE) catalog. The official government spatial reference mask and tabular wheat-area statistics used for validation are third-party data subject to institutional restrictions. However, the machine learning models, XGBoost inference pipeline, and curated spatial datasets developed during this study are publicly available for exploration and reproducibility on Hugging Face.\footnote{Code and trained models are publicly available at: \url{https://huggingface.co/spaces/kasimali/kasimali}}

\newpage
\appendix
\renewcommand{\thefigure}{A\arabic{figure}}
\setcounter{figure}{0}
\section*{Appendix: Supplementary Material}
\label{sec:appendix}

\begin{figure}[H]
\centering
\includegraphics[width=\textwidth, keepaspectratio]{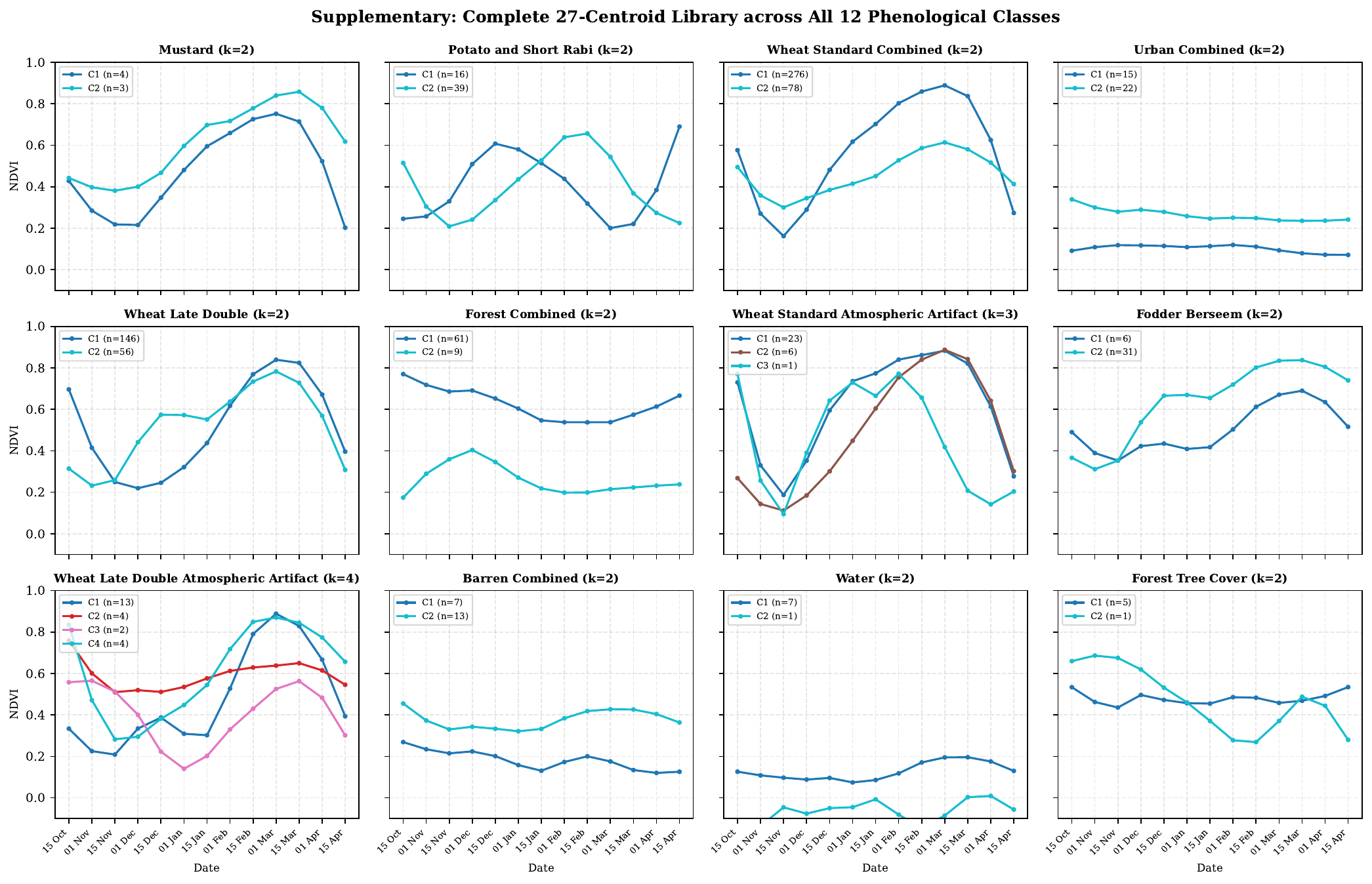}
\caption{Centroid library computed by $k$-means clustering on the final 849-sample dataset, shown for display: all 27 sub-cluster phenological curves across the twelve master classes. This display library is distinct from the iterative libraries used by the rule engine (Section~\ref{sec:batch2}).}
\label{fig:fullgrid}
\end{figure}

Figure~\ref{fig:fullgrid} presents the centroid library computed on the final 849-sample dataset (Section~\ref{sec:batch2}): all 27 sub-cluster phenological curves across the twelve master classes, provided in full beyond the four wheat-related classes highlighted in Figure~\ref{fig:centroidcomp}.

\end{document}